\pdfoutput=1
\documentclass[11pt]{article}

\usepackage[preprint]{acl}

\usepackage{times}
\usepackage{latexsym}
\usepackage[T1]{fontenc}
\usepackage[utf8]{inputenc}
\usepackage{microtype}
\usepackage{inconsolata}
\usepackage{tcolorbox}
\usepackage{graphicx}
\graphicspath{{image/}}
\usepackage{pdfpages}
\usepackage{url}
\usepackage{hyperref}
\usepackage{xcolor}
\usepackage{epsfig}
\usepackage{adjustbox}
\usepackage{amsfonts}
\usepackage{amsmath}
\usepackage{amssymb}
\usepackage{booktabs} 
\usepackage{comment}
\usepackage{multirow}
\usepackage{caption}
\usepackage{subcaption}
\usepackage{textcomp}
\usepackage{relsize}
\usepackage{stmaryrd}
\usepackage{bbm}
\usepackage[htt]{hyphenat}
\usepackage{soul}
\usepackage{float}
\usepackage{hyperref}
\usepackage{placeins}
\usepackage{tablefootnote}

\usepackage{xspace}
\newcommand{\ours}[0]{\textsc{NarrativeFactScore}\xspace}
\newcommand{\oursshort}[0]{\textsc{NFS}\xspace}

\title{Agent-as-Judge for Factual  Summarization of Long Narratives}

\author{
    Yeonseok Jeong\textsuperscript{1},
    Minsoo Kim\textsuperscript{1}, 
    Seung-won Hwang\textsuperscript{2}\thanks{Corresponding Author},
    Byung-Hak Kim\textsuperscript{3} \\
    IPAI, Seoul National University\textsuperscript{1}, 
    Seoul National University\textsuperscript{2},
    Hyundai Card\textsuperscript{3} \\
    \texttt{\{jys3136, minsoo9574, seungwonh\}@snu.ac.kr} \\
    \texttt{byunghak.kim@hcs.com}
}

\begin{document}
\maketitle

\begin{abstract}
Large Language Models (LLMs) have demonstrated near-human performance in summarization tasks based on traditional metrics such as ROUGE and BERTScore. 
However, these metrics do not adequately capture critical aspects of summarization quality, such as factual accuracy, particularly for long narratives (>100K tokens). 
Recent advances, such as \textit{LLM-as-a-Judge}, address the limitations of metrics based on lexical similarity but still exhibit factual inconsistencies, especially in understanding character relationships and states.
In this work, we introduce \ours (\oursshort), the first ``Agent-as-a-Judge'' framework that evaluates and refines factuality in narrative summarization.
By leveraging a Character Knowledge Graph (CKG) extracted from input narrative, \ours evaluates the factuality and provides actionable guidance for refinement, such as identifying missing or erroneous facts. 
Our experimental results demonstrate that constructing the CKG enables reasoning with 1/3 of the factuality computation used in the prior approach, and achieve three times higher correlation with human judgments.
Furthermore, refinement with actionable guidance improves the quality of the summary.\footnote{\href{https://github.com/YeonseokJeong/NarrativeFactScore}{https://github.com/YeonseokJeong/NarrativeFactScore}}
\end{abstract}

\section{Introduction}
\label{sec:introduction}
\begin{figure}[h]
{
\centering
    \includegraphics[width=1.0\linewidth]{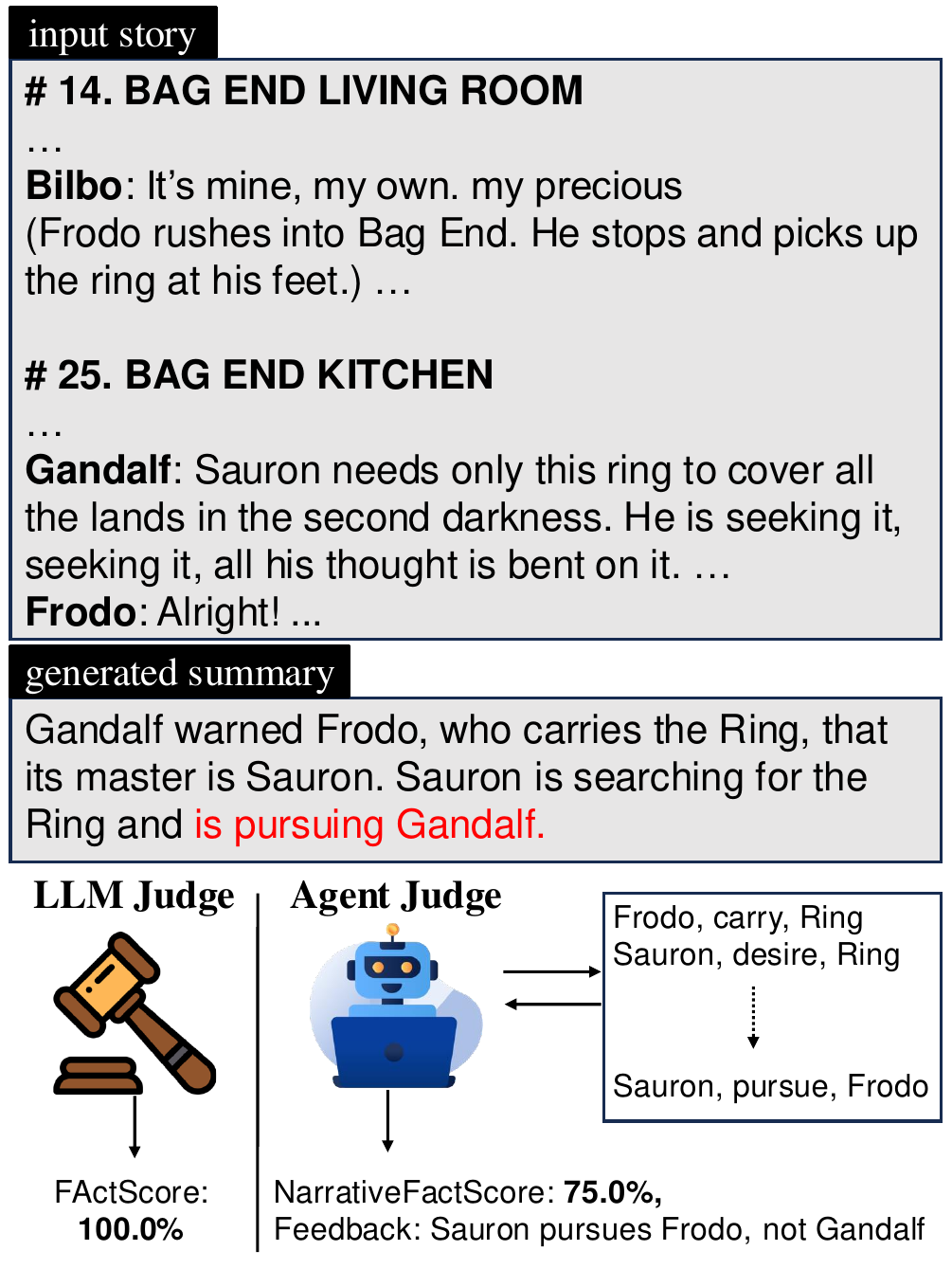}
    \vspace{-7mm}
    \caption{
    Comparison of factuality evaluation by LLM and Agent Judge with \ours. Given scenes from \textit{The Lord of the Rings}, the summary incorrectly claims ``Sauron is pursuing Gandalf.'' The LLM Judge assigns 100\% factuality score, while our Agent Judge correctly identifies this error through analyzing atomic facts about characters, assigning 75\% \ours, with specific feedback.
    \label{fig:limitation_example}
    }
}
\vspace{-5mm}
\end{figure}

The rise of LLMs~\cite{openai2023gpt4,dubey2024llama} has brought significant advancements to summarization tasks, achieving performance close to human levels~\cite{pu2023summarization}.
Most evaluation metrics~\cite{lin2004rouge, zhang2019bertscore, yuan2021bartscore} for summarization measure lexical or semantic similarity between summary and ground truth.

In our target scenario of summarizing long narratives (> 100K tokens), metrics such as BooookScore~\cite{chang2024booookscore} can measure coherence, but evaluating factuality has remained challenging~\cite{subbiah2024storysumm}.
This is because it requires comparing summaries not only against complex facts but also against the evolving relationships among characters in long narratives.
Thus, judging the factuality of such long narratives has therefore inevitably relied on costly human evaluations~\cite{kim2024fables}.

More recently, \textit{LLM-as-a-Judge} metrics~\cite{min-etal-2023-factscore, bishop-etal-2024-longdocfactscore} have leveraged LLM to assess the factuality, offering a more cost-effective alternative to human annotations.
If applied to narrative summarization, these metrics split the summary into smaller units, retrieve similar scenes from the input story, and quantify factuality by LLM.

However, directly using LLM to evaluate factuality has two limitations.
\textbf{First,} as demonstrated by \citet{kim2024fables}, the LLM judge often fails to accurately assess factuality in narratives that require indirect \textbf{reasoning}, such as understanding character relationships or states.
For example, in Figure~\ref{fig:limitation_example}, although Sauron is pursuing Frodo in order to obtain the Ring in \texttt{The Lord of the Rings}, the LLM judge inaccurately evaluates the factuality of a summary which incorrectly reports that ``Sauron is pursuing Gandalf''.
This limitation stems from the inability of the LLM judge to consistently track and reason about character relationships.
To address this, we introduce the CKG, a structured representation of characters and their relationships that must be maintained consistently.

\textbf{Second,} the LLM judge outputs only a single score with limited \textbf{interpretability}, which makes it less reliable and difficult to identify what needs to be improved.
Desirably, evaluation metrics for summarization can provide feedback, when the score is low, to explain why it is incorrect and suggest how to improve.

We propose an \textit{Agent-as-a-Judge}~\cite{zhuge2024agent} framework, using interpretable evaluation of summaries with a novel \textbf{\ours}, based on which we can refine and improve summary quality.
CKG achieves consistency by constructing a names graph that consolidates character aliases and variations across scenes and by performing multiple rounds of relationship extraction, selecting relationships that frequently appear across scenes as edges, inspired by \citet{wang2023selfconsistency}.
This construction process ensures that only well-supported character relationships are retained.
By leveraging this consistent relationship graph when evaluating the factuality, we can accurately assess even complex narrative facts that require understanding intricate character dynamics.

To improve the interpretability of the metric, \ours also provides feedback for interpretation and refinement when the summary is incorrect.
For each statement in the summary, our metric retrieves relevant scenes and character relationships from our CKG to calculate a factuality score. 
Based on the retrieved evidence, ours evaluates each statement and generates feedback identifying discrepancies between claims and supporting evidence.
Since our metric operates autonomously, it is more cost-effective and faster than \textit{Human-as-a-Judge}.
In addition, it offers feedback for low scores, which makes it more interpretable than \textit{LLM-as-a-Judge} metrics.
Recognizing the causes of low scores also contributes to generating more accurate summaries through agent-based refinement.

Using \ours provides two key advantages for long narrative summarization.
First, it offers a labor-efficient and fast metric that also approximates human evaluation when evaluating the factuality of summaries.
Our metric demonstrates a statistically strong correlation with human evaluation, and a test for differences between human evaluation and our metric yielded statistically significant results, with the p-value falling below 0.05.
Second, by providing feedback on factually incorrect parts, it can improve summarization performance.
We show that agent-based refinement improves factuality (+14.03), ROUGE (+2.05), and BERTScore (+0.13) on MovieSum~\cite{saxena2024moviesum}, a movie script summarization dataset, and also improves factuality (+12.26), ROUGE (+2.47), and BERTScore (+0.21) on MENSA~\cite{saxena2024select}, a movie scene saliency dataset.
\section{Related Work}
\label{sec:related_work}
\subsection{Long Narrative Summarization}
Summarizing long narratives \cite{saxena2024moviesum, saxena2024select} is challenging due to the high computational and memory demands required by transformer-based models.
In prior work \cite{pilault-etal-2020-extractive, li2021ease, wu2021recursively, chang2024booookscore}, a method called \textit{hierarchical merging} was introduced, where individual chunks of the narrative are summarized separately and then combined to form a coherent final summary.
Other segmentation-aware strategies have also been explored \cite{moro2022semantic_self_segmentation, moro2023align_then_abstract, zhang2022summn}.
But such chunk-based methods can group unrelated content into the same segment.
Although these methods preserve the logical structure of the narrative, hallucinations remain a frequent challenge, especially when capturing global information such as character relationships.
Thus, our focus is on improving the factuality of the summaries.

\subsection{Character Knowledge Graph (CKG)}
Since characters are integral to narrative~\cite{gurung-lapata-2024-chiron}, prior work has aimed to construct a graph to easily utilize them.
In narrative texts, CKG shows the unidirectional relationship between a subject and an object character.
This process is similar to creating a triple (subject-predicate-object) list in knowledge graph construction~\cite{chen2020knowledge}.
\citet{andrus2022enhanced} utilized the OpenIE system \cite{angeli-etal-2015-leveraging} for story completion and question-answering tasks, integrating it with GPT-3 \cite{brown2020language} to enhance its effectiveness.
Alternatively, a recent method \cite{zhao-etal-2024-large} that assembles CKG directly using LLMs is a more robust approach, as it better captures the nuanced and complex relationships.
Our distinction lies not only in constructing CKGs but also in utilizing them to measure and enhance factuality.

\subsection{Summarization Metrics for Evaluating Factuality}
In recent research, efforts have been made to evaluate factuality of long documents.
LongDocFACTScore \cite{bishop-etal-2024-longdocfactscore} improves this process by calculating BARTScore~\cite{yuan2021bartscore} only on the semantically similar portions of the source text for each summary sentence, making it an effective method for handling long documents.
FActScore \cite{min-etal-2023-factscore} further enhances factuality evaluation by decomposing text into atomic facts and verifying each with LLM using information retrieved from the knowledge source.
Unlike these metrics, our metric focuses on character relationships to accurately evaluate factuality and provide actionable feedback to refine factually incorrect parts.
\section{Proposed Method}
\label{sec:3}
\begin{figure*}[!t]
 	\centering
 	\includegraphics[width=0.99\linewidth]{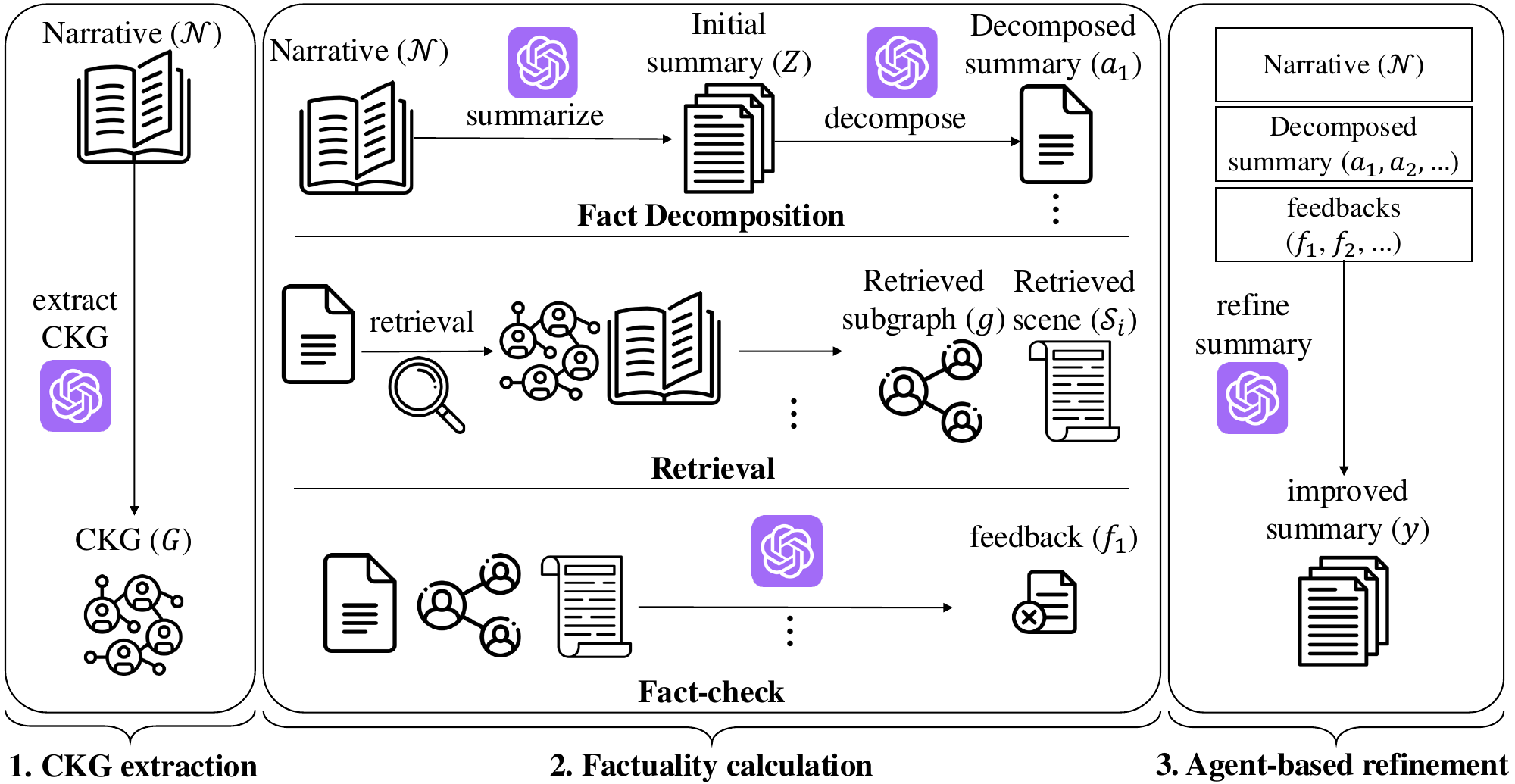}
 	\caption{The main figure illustrates the overall process of evaluation and refinement, which includes three main stages. 
  First, it shows the extraction of CKG $G$ from narrative $\mathcal{N}$.
  Next, it depicts the calculation of factuality by comparing the decomposed summary $a_k$ against the retrieved character relationship subgraph $g$ and narrative scene $\mathcal{S}_i$.
  Finally, it illustrates the agent-based refinement process, where feedbacks ($f_1, f_2, ...$) are used to improve the factual accuracy of the summary.
  }
 	\label{fig:main_method}
\end{figure*}
In this section, we elaborate on \ours for evaluating factuality of long narrative summarization.
Figure~\ref{fig:main_method} illustrates three phases of our framework, which will be detailed in Section~\ref{sec:3_1}, \ref{sec:3_2}, and \ref{sec:3_3} respectively.

\subsection{CKG Extraction}
\label{sec:3_1}
We construct a \textbf{consistent} CKG, to overcome the inconsistencies of CKG reported in \citet{kim2024fables, zhao-etal-2024-large}, losing information~\cite{liu2024lost} in long narratives and failing to reason over many implicit relationships at once.
To address these issues, we perform reasoning multiple times~\cite{wang2023selfconsistency} for each scene and select frequent relationships to improve consistency and accuracy.
We note this requires only 1/3 of the original cost while improving correlation threefold. (See Appendix~\ref{sec:analysis_latency}.)

Given a narrative represented as a collection of scenes $\mathcal{N} = \{\mathcal{S}_1, \mathcal{S}_2, \ldots, \mathcal{S}_m\}$, where $m$ denotes the number of scenes, the goal is to extract a graph $G$ that encapsulates character relationships. 
Each scene $\mathcal{S}_i\ (1 \leq i \leq m)$ is processed individually to extract relation triples (subject-predicate-object) using GPT-4o-mini~\cite{openai2023gpt4}, as detailed in Section~\ref{sec:prompt_knowledge_extraction}.
The extracted triples are used to initialize the nodes and determine the edges based on the main relationships between the nodes, forming the final CKG $G$ through the following two steps.

\begin{figure*}[t!]
\begin{subfigure}{0.5\textwidth}
    \includegraphics[width=\linewidth]{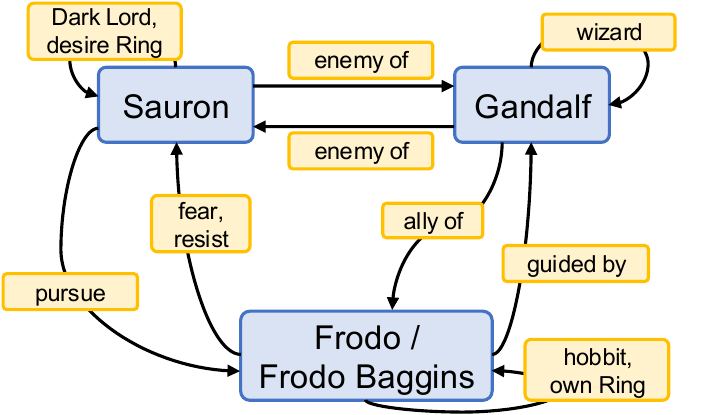}
    \vspace{13pt}
    \caption{Knowledge graph.} \label{fig:knowledge_graph}
\end{subfigure}
\hfill
\begin{subfigure}{0.35\textwidth}\small
    <subject>Frodo \\
    \null\qquad<predicate>hobbit, own Ring \\
    \null\quad<object>Gandalf \\
    \null\qquad<predicate>guided by \\
    \null\quad<object>Sauron \\
    \null\qquad<predicate>fear, resist \\
    <subject>Gandalf \\
    \null\qquad<predicate>wizard \\
    \null\quad<object>Frodo \\
    \null\qquad<predicate>ally of \\
    \null\quad<object>Sauron \\
    \null\qquad<predicate>enemy of \\
    <subject>Sauron \\
    \null\qquad<predicate>Dark Lord, desire Ring \\
    \null\quad<object>Frodo \\
    \null\qquad<predicate>pursue \\
    \null\quad<object>Gandalf \\
    \null\qquad<predicate>enemy of \\
    \caption{Linearized knowledge graph.} \label{fig:linearized_knowledge_graph}
\end{subfigure}

\caption{(a) Part of a knowledge graph generated from \textit{The Lord of the Rings}, with three named entities. `Frodo/Frodo Baggins' is a single entity with two names. (b) The same graph is in linearized form.}
\vspace{-5mm}
\end{figure*}
 
First, to maintain consistency in character identification, we construct a \textbf{names graph} $G_{name}$, consolidating aliases or variations in names in scenes. 
Our framework processes each scene in turn, extracts all character names, and determines several names refer to the same character based on the context using LLM.
For example, in \texttt{The Lord of the Rings}, `Frodo' and `Frodo Baggins' are recognized as the same character. 
As illustrated by `Frodo / Frodo Baggins' in Figure~\ref{fig:knowledge_graph}, each name variation is a node.\footnote{In practice, an undirected edge is added between nodes that refer to the same character.}
This step ensures an accurate capture of relationships, even when names vary across scenes.
The CKG is initialized using names from the names graph.

Second, to preserve the consistency of relationships, we sample extracted triples multiple times~\cite{wang2023selfconsistency, brown2024large} and select frequent ones as the final edges.
Let the node set $V$ be the set of all characters in the names graph:
\begin{equation}
V = \{ v \mid v \in G_{\text{name}} \}
\end{equation}
Only triples with named entities as subjects and objects are used; if an object is missing, a self-loop is added to represent the state of a character.
We then define the edge set $E$ of our CKG as 
\begin{equation}
E = \bigl\{(s,p,o)\mid s,o\in V,\ \mathrm{freq}(p\mid s,o)\ge\tau \bigr\}
\label{eq:edge_def}
\end{equation}
where $(s,o)$ denotes a character pair, $\mathrm{freq}(p\mid s,o)$ is the frequency of predicate $p$ for $(s,o)$, and $\tau$ is the frequency threshold.\footnote{Adjusting the threshold allows for control over the graph: a higher threshold ensures greater consistency, while a lower threshold increases diversity.}
Finally, the consistent CKG is given by
\begin{equation}
G = (V, E).
\end{equation}

For triples with the same subject and object, frequent predicates capture temporal changes as directed edges.
For triples with the same subject and object, our CKG saves predicates that occur frequently across multiple scenes.
By storing these frequent predicates in sequential order, our framework captures temporal narrative shifts as directed edges (Eq~\ref{eq:edge_def}).
For example, since the early scenes show that `Frodo' fears `Sauron' and the later scenes show that he resists `Sauron', the CKG in Figure~\ref{fig:knowledge_graph} displays two distinct relationships.
The process of deciding edges is repeated to construct a CKG that can effectively evaluate the factuality of summaries.

\subsection{\ours Calculation}
\label{sec:3_2}
We invent a new metric to guide agentic evaluation, unlike existing factuality metrics~\cite{min-etal-2023-factscore, bishop-etal-2024-longdocfactscore} that do not provide evidence or feedback for their scores, by considering events in the input story superficially but overlooking relational information about characters.
Our metric addresses these limitations by incorporating character relationship graphs and providing detailed feedback.
To calculate the factuality of the narrative summary, we first generate an initial summary $Z$ using the prompt described in Section~\ref{sec:prompt_initial_summarization}.

To evaluate the factuality of the initial summary $Z$, we decompose it into smaller verifiable units, similar to the approach used in \citet{min-etal-2023-factscore}.
Using the prompt in Section~\ref{sec:prompt_fact_decompose}, each sentence in the initial summary $Z$ is divided into a list of atomic facts $A = \{a_1, a_2, \ldots, a_z\}$.

To evaluate each atomic fact $a_k$, we need the scene and information about the characters that appear in the atomic fact.
First, we retrieve the most relevant scene $\mathcal{S}_i$ within the narrative $\mathcal{N}$, by using the BGE-M3 \cite{chen2024bge}.
Second, we also retrieve the subgraph $g$ from the linearized CKG $G$, as illustrated in Figure \ref{fig:linearized_knowledge_graph}, by sorting candidate triples by similarity to the atomic fact $a_k$.

Using the retrieved information, each atomic fact $a_k$ is evaluated to determine its factuality and to obtain feedback supporting the evaluation.
We then define an indicator $I_k$ for factuality based on:
\begin{equation}
I_k = \mathbbm{1}[a_k \ \text{is factual given} \ \mathcal{S}_i, g)]
\end{equation}
where $\mathbbm{1}$ is the indicator function, yielding 1 if the atomic fact $a_i$ is factual and 0 otherwise.
This evaluation is carried out using the prompt detailed in Section~\ref{sec:prompt_fact_check}, which produces 1 if the atomic fact is accurate.
If the atomic fact is determined to be inaccurate, then feedback $f_k$ is provided on how to correct it. 
Finally, the \ours is calculated as the proportion of atomic facts that are found to be factual, defined by the following equation:
\begin{equation}
\text{\ours} = \frac{\sum_{i=1}^{z} I_k}{z}
\end{equation}

\subsection{Agent-based Fact Refinement}
\label{sec:3_3}
The new metric leveraging consistent CKG enables the LLM agent to guide refinement by using feedback from the evaluation.
This process involves three key inputs: original narrative to provide global context, the initial summary that requires modification, and the feedback detailing the inaccuracies and reasons for those errors.
Using a prompt that incorporates these inputs in Section~\ref{sec:prompt_self_improvement},  LLM refines the initial summary by correcting only the factually inaccurate parts provided from feedback, then generates the improved summary $y$.

Motivated by \citet{madaan2024self}, the improved summary can be further evaluated as outlined in Section~\ref{sec:3_2}. 
This allows the agent-based refinement to be iterative, where each iteration further refines the summary to enhance overall factuality.
\section{Experiments}
\subsection{Implementation Details}
\paragraph{Uniform Language Model Usage}
To ensure that performance gain is due to our framework and not the underlying language model, we use only \texttt{gpt-4o-mini-2024-07-18} in our experiments. 
This model is applied across all components, including CKG extraction, summarization, fact decomposition, fact check, and agent-based fact refinement. 
This approach prevents superior LLMs from influencing the results, allowing us to rigorously evaluate the effectiveness of our framework.

\paragraph{Generating Initial Summary}
To generate the initial summary $Z$, we adopt \textit{hierarchical merging}~\cite{chang2024booookscore} that ensures the logical structure of the narrative is preserved. 
The narrative is first divided into chunks $C_i$ where each chunk is formed incrementally by adding scenes until a predefined context size\footnote{we set the predefined context size of a chunk to 1024.} is reached. 
Once this limit is exceeded, a new chunk begins, resulting in a sequence of chunks $\mathcal{C} = \{C_1, C_2, ..., C_n\}$.
Each chunk $C_i$ is then independently summarized using the prompt specified in Section~\ref{sec:prompt_initial_summarization}, and the resulting chunk summaries are sequentially merged to produce the initial summary $Z$.

\paragraph{Retrieving Relevant Scene and Subgraph}
Using the BGE-M3 embedding model \cite{chen2024bge}, we retrieve information relevant to each atomic fact $a_k$.
Specifically, we identify the most similar scene $\mathcal{S}_i$ from the narrative $\mathcal{N}$ and a subgraph containing the three most relevant triples in the linearized CKG $G$.
All retrieval computations were performed on a single NVIDIA RTX 3090 GPU.

\subsection{Evaluation Metrics}
We assess the performance of our framework using several key evaluation metrics. 
ROUGE~\cite{lin2004rouge} assesses n-gram overlap with reference summaries, including R-1 (unigram), R-2 (bigram) and R-L (longest common subsequence).
BERTScore~\cite{zhang2019bertscore} (BS$\mathbf{_p}$, BS$\mathbf{_r}$, BS$\mathbf{_{f1}}$) evaluates similarity using BERT embeddings~\cite{devlin-etal-2019-bert}, where BS$\mathbf{_p}$ represents precision, BS$\mathbf{_r}$ recall and BS$\mathbf{_{f1}}$ the F1-score. 
BARTScore~\cite{yuan2021bartscore} measures the quality of summaries by scoring them as conditional language generation tasks.
Finally, we propose \ours (\oursshort) as a novel metric to measure the factuality of the generated summaries.

We report ROUGE and BERTScore as reference points for lexical and semantic similarity, while emphasizing that these metrics were not designed to capture factual accuracy. 
Our primary factuality comparisons are instead carried out with dedicated metrics such as FActScore~\cite{min-etal-2023-factscore} and LongDocFACTScore~\cite{bishop-etal-2024-longdocfactscore}.

\subsection{Correlation with Human Factuality Scores}
\begin{table}[h!]
\centering
\resizebox{1.0\linewidth}{!}{
\begin{tabular}{l|ccc}
\noalign{\hrule height 1pt}
\textbf{Dataset} & \textbf{\# tokens (source)} & \textbf{\# scenes} & \textbf{\# tokens (summary)} \\
\hline
FABLES & 127,467 & 38 & 594 \\
STORYSUMM & 782 & -- & 322 \\
\noalign{\hrule height 1pt}
\end{tabular}
}
\caption{Dataset statistics for STORYSUMM and FABLES.}
\label{tab:dataset_stats}
\end{table}

\begin{table}[ht]
\centering
\resizebox{\linewidth}{!}{%
\begin{tabular}{l|cc|cc}
\noalign{\hrule height 1pt}
\multirow{2}{*}{\textbf{Metrics}} & \multicolumn{2}{c|}{\textbf{STORYSUMM}} & \multicolumn{2}{c}{\textbf{FABLES}} \\
 & \textbf{Spearman} & \textbf{KENDALL} & \textbf{Spearman} & \textbf{KENDALL} \\
\hline
ROUGE-1 & 0.25 & 0.18 & -0.20 & -0.14 \\
ROUGE-2 & 0.30 & 0.22 & -0.04 & -0.03 \\
ROUGE-L & 0.31 & 0.22 & -0.18 & -0.14 \\
BERTScore$\mathbf{_{f1}}$ & 0.19 & 0.13 & -0.13 & -0.08 \\
BARTScore & 0.09 & 0.06 & -0.30 & -0.22 \\
LongDocFACTScore  & 0.07 & 0.05 & 0.24 & 0.16 \\
FActScore & 0.19 & 0.13 & 0.16 & 0.09 \\
\textsc{\oursshort} & \underline{\textbf{0.43}} & \underline{\textbf{0.31}} & \underline{\textbf{0.47}} & \underline{\textbf{0.33}} \\
\noalign{\hrule height 1pt}
\end{tabular}%
}
\caption{Spearman and KENDALL's tau correlation coefficients between different metrics and human factuality assessments on STORYSUMM and FABLES. Coefficients indicating strong correlation are \underline{underlined}.\tablefootnote{We follow widely adopted interpretations reported in Table~\ref{tab:coefficient}.}}
\label{tab:correlation}
\end{table}
\begin{table}[ht]
\centering
\resizebox{\linewidth}{!}{%
\begin{tabular}{l|cc|cc}
\noalign{\hrule height 1pt}
\multirow{2}{*}{\textbf{Metrics}} & \multicolumn{2}{c|}{\textbf{STORYSUMM}} & \multicolumn{2}{c}{\textbf{FABLES}} \\
 & \textbf{Spearman} & \textbf{KENDALL} & \textbf{Spearman} & \textbf{KENDALL} \\
\hline
(A) \oursshort & \textbf{0.43} & \textbf{0.31} & \textbf{0.47} & \textbf{0.33} \\
(B) \phantom{0}$-$ consistency & 0.21 & 0.14 & 0.19 & 0.13 \\
(C) \phantom{0}$-$ CKG & 0.30 & 0.21 & 0.25 & 0.16 \\
\noalign{\hrule height 1pt}
\end{tabular}
}
\caption{Ablation results on STORYSUMM and FABLES, showing the impact of using different CKG.}
\label{tab:ablation}
\end{table}

\paragraph{Dataset} 
To evaluate whether the \ours we proposed correlates effectively with human factuality, we required benchmarks that satisfy two conditions.
First, they should provide human factuality scores for multiple LLM-generated summaries of each narrative, enabling correlation analysis.
Second, they should be released after October 2023 to minimize potential contamination given the GPT-4o-mini knowledge cutoff.
STORYSUMM~\cite{subbiah2024storysumm} and FABLES~\cite{kim2024fables} are the only datasets that meet both conditions, so we chose them as our benchmarks in Tables~\ref{tab:correlation} and \ref{tab:ablation}.

To further explain these datasets, Table~\ref{tab:dataset_stats} shows their statistics. 
STORYSUMM is relatively short (782 tokens on average, 322 in summaries) and thus scenes were not separated, while FABLES is substantially longer (>100K tokens with 38 scenes). 
These statistics confirm that our method effectively calculates factuality for both short- and long-length narratives.

\paragraph{Results}
We computed the Spearman~\cite{spearman1961proof} correlations and KENDALL's tau~\cite{kendall1938new} correlations for each metric in relation to the human factuality scores, as shown in Table \ref{tab:correlation}.
\ours is the only metric that shows a strong correlation with human annotations in all datasets.
This correlation is statistically significant, with p-values below 0.05 for all datasets.

\paragraph{Ablation Study}
To verify the effectiveness of CKG in evaluating factuality, we conducted an ablation study in Table~\ref{tab:ablation}.
\ours (A) iteratively reasons about character relationships, selects frequent relationships to construct a consistent CKG, and utilizes it for factuality evaluation.
In contrast, (B) generates the CKG by reasoning character relationships in a single step and evaluates factuality accordingly.
However, according to \citet{zhao-etal-2024-large}, LLMs tend to generate inaccurate character relationships when reasoning over long narrative in a single step.
Lastly, (C) evaluates factuality without utilizing a CKG.

The experimental results show that our metric (A) achieves the highest correlation with human, and indicate the following observations.
First, comparing (A) with (C) shows CKG contributes to more accurate factuality evaluation.
However, the results of (B) and (C) show that an inaccurate CKG can hinder factuality evaluation rather than improve it.
Thus, to effectively assess the factuality of summary, it is necessary to construct a consistent CKG through multiple iterations of reasoning.
\begin{table*}[h!]
\centering
\resizebox{1.0\textwidth}{!}{
\begin{tabular}{l|ccccccc|ccccccc}
\noalign{\hrule height 1pt} 
\textbf{} & \multicolumn{7}{c|}{\textbf{MENSA}} & \multicolumn{7}{c}{\textbf{MovieSum}} \\
\textbf{} & \textbf{R-1} & \textbf{R-2} & \textbf{R-L} & \textbf{BS$\mathbf{_p}$} & \textbf{BS$\mathbf{_r}$} & \textbf{BS$\mathbf{_{f1}}$} & \textbf{\oursshort} & \textbf{R-1} & \textbf{R-2} & \textbf{R-L} & \textbf{BS$\mathbf{_p}$} & \textbf{BS$\mathbf{_r}$} & \textbf{BS$\mathbf{_{f1}}$} & \textbf{\oursshort} \\
\hline
\textbf{\textit{without merging}} & \multicolumn{7}{c|}{} & \multicolumn{7}{c}{} \\
~~ TextRank & \textbf{34.37} & 4.60 & 12.84 & 46.86 & 49.43 & 48.10 & 59.72 & \textbf{33.92} & 4.62 & 16.25 & 46.82 & 49.48 & 48.10 & 60.23 \\
~~ LED & 17.46 & 1.59 & 10.03 & 42.90 & 42.74 & 42.58 & 56.48 & 2.80 & 0.28 & 0.28 & 32.64 & 23.82 & 27.32 & 22.24 \\
~~ LongT5 & 20.77 & 2.26 & 10.03 & 45.05 & 45.06 & 45.01 & 73.76 & 20.18 & 1.99 & 13.83 & 44.58 & 44.28 & 44.36 & 74.01 \\
\hline
\textbf{\textit{hierarchically merging}} & \multicolumn{7}{c|}{} & \multicolumn{7}{c}{} \\
~~ GPT-4o-mini & 31.79 & 9.69 & 12.68 & 60.00 & 60.03 & 60.01 & 81.05 & 29.26 & 8.72 & 17.88 & 59.11 & 59.29 & 59.19 & 80.56 \\
~~  Ours: 1st iteration & 33.00 & 9.70 & 12.84 & 60.22 & 60.11 & 60.16 & 85.94 & 30.36 & 8.74 & 18.55 & 59.26 & 59.30 & 59.27 & 86.92 \\
~~  Ours: 2nd iteration & 33.75 & 9.72 & 13.07 & 60.17 & 60.10 & 60.12 & 88.94 & 30.98 & 8.75 & 18.61 & 59.33 & 59.30 & 59.30 & 92.04 \\
~~  Ours: 3rd iteration & 34.26 & \textbf{9.74} & \textbf{13.46} & \textbf{60.24} & \textbf{60.21} & \textbf{60.22} & \textbf{93.31} & 31.31 & \textbf{8.81} & \textbf{18.62} & \textbf{59.36} & \textbf{59.31} & \textbf{59.32} & \textbf{94.59} \\
\noalign{\hrule height 1pt}
\end{tabular}
}
\caption{Evaluation results on MENSA \cite{saxena2024select} and MovieSum \cite{saxena2024moviesum} datasets.}
\label{tab:main_result}
\end{table*}

\subsection{Summarization Performance Evaluation}
\paragraph{Datasets}
We evaluated our framework on the MENSA \cite{saxena2024select} and MovieSum \cite{saxena2024moviesum} datasets. 
MENSA aligns movie scenes with Wikipedia summaries and MovieSum pairs screenplays with summaries.
Since both datasets provide ground-truth summaries, they can be used to test whether our agent-based refinement improves both factuality and summarization quality.
We use the full test sets: 50 samples from MENSA and 200 from MovieSum.

\paragraph{Results}
We evaluated summarization performance using two baseline types. 
The first type includes methods \textit{without merging} that summarize all input in a single step, such as TextRank~\cite{mihalcea2004textrank}, Longformer Encoder-Decoder (LED)~\cite{beltagy2020longformer}, and LongT5~\cite{guo2022longt5}.
The second type involves \textit{hierarchical merging}~\cite{chang2024booookscore}, with which we performed experiments using GPT-4o-mini~\cite{openai2023gpt4}.
Additionally, we evaluated the summaries generated by GPT-4o-mini after agent-based iterative refinements (1st to 3rd).

As shown in Table~\ref{tab:main_result}, agent-based refinement improves not only factuality but also other metrics, improving the overall quality of the summaries.
This refinement improves performance consistently, yielding improvements of +14.03 in factuality, +2.05 in ROUGE, and +0.13 in BERTScore on MovieSum, and +12.26, +2.47, and +0.21 respectively on MENSA.

\section{Analysis}
\label{sec:analysis}
\subsection{How Consistently Does Ours Capture Character Relationships?}
\begin{table}[]
\centering
\resizebox{1.0\linewidth}{!}{
\begin{tabular}{lccc}
\noalign{\hrule height 1pt}
\textbf{Method} & \textbf{BS$\mathbf{_p}$} & \textbf{BS$\mathbf{_r}$} & \textbf{BS$\mathbf{_{f1}}$} \\
\hline
Naive extract~\cite{zhao-etal-2024-large}   & 86.23          & 86.33          & 86.26           \\
Ours            & \textbf{95.63}          & \textbf{95.68}          & \textbf{95.65}           \\
\noalign{\hrule height 1pt}
\end{tabular}}
\caption{Comparison between the naive extract method and our proposed method.}
\label{tab:kg_accuracy}
\end{table}
To effectively evaluate factuality and improve summary, it is necessary to generate an accurate and consistent CKG.
According to \citet{kim2024fables, zhao-etal-2024-large}, the ``naive extract'' approach, where an LLM extracts character relationships in one step, often fails to consistently capture some relationships.
Thus, our objective is to verify whether our approach can generate a consistent CKG.
Unlike other datasets, Conan~\cite{zhao-etal-2024-large} provides ground truth annotation of character relationships within narratives.
To evaluate whether the generated relation is semantically similar to this ground truth, we measure the BERTScore~\cite{devlin-etal-2019-bert}.

As shown in Table~\ref{tab:kg_accuracy}, our method generates CKG that is closely similar to ground truth.
Although the ``naive extract'' achieves 86.26, it occasionally produces incorrect relationships.
In contrast, by reasoning about relationships scene by scene and aggregating them, our method chooses more accurate relationships and constructs a consistent CKG.
This supports the role of the CKG in enhancing factuality assessment, as shown in Table~\ref{tab:ablation}.
Even without explicit scene boundaries, as in the Conan dataset, we segmented the text into 256-token chunks with 128-token overlaps.
This approach still showed a high BERTScore, demonstrating robustness.

\subsection{Challenging Set}
\begin{table}[h!]
\centering
\resizebox{1.0\linewidth}{!}{
\begin{tabular}{l|ccccccc}
\noalign{\hrule height 1pt}
\textbf{} & \textbf{R-1} & \textbf{R-2} & \textbf{R-L} & \textbf{BS$\mathbf{_p}$} & \textbf{BS$\mathbf{_r}$} & \textbf{BS$\mathbf{_{f1}}$} & \textbf{\oursshort} \\
\hline
\textbf{\textit{without merging}} & \multicolumn{7}{c}{} \\
~~ TextRank & \textbf{33.92} & 4.63 & 16.25 & 46.82 & 49.48 & 48.10 & 62.43 \\
~~ LED & 2.75 & 0.17 & 0.64 & 31.78 & 24.44 & 27.37 & 11.70 \\
~~ LongT5 & 22.10 & 2.29 & 11.16 & 43.86 & 44.69 & 44.18 & 79.38 \\
\hline
\textbf{\textit{hierarchically merging}} & \multicolumn{7}{c}{} \\
~~ GPT-4o-mini & 28.07 & 8.01 & 14.12 & 58.37 & 59.36 & 58.53 & 81.30 \\
~~ Ours: 1st iteration & 29.02 & 8.09 & 14.08 & 58.49 & 59.30 & 58.86 & 84.59 \\
~~ Ours: 2nd iteration & 29.98 & 8.19 & 14.24 & 58.61 & 59.32 & 58.94 & 90.47 \\
~~ Ours: 3rd iteration & 30.22 & \textbf{8.20} & \textbf{14.44} & \textbf{58.75} & \textbf{59.39} & \textbf{59.04} & \textbf{93.22} \\
\noalign{\hrule height 1pt}
\end{tabular}
}
\caption{Evaluation results on the challenging set of the MovieSum~\cite{saxena2024moviesum} dataset.}
\label{tab:challeng_result}
\end{table}

We aim to evaluate whether our metric can provide feedback necessary to improve factuality in recent narratives.
Although LLM-based metrics provide accurate factuality feedback for narratives within their pretraining data, they tend to be less reliable for narratives outside of their training corpus.
However, our metric provides accurate feedback by evaluating summaries based on narrative story and character relationships rather than relying on parametric knowledge alone.
Therefore, we define a challenging set of works published after the knowledge cutoff date of our LLM to verify whether our metric improves factuality through its feedback.

Our metric demonstrates the capability to provide feedback for improving factuality even in recent works.
For this experiment, we curated a challenging set of 18 movies from MovieSum~\cite{saxena2024moviesum} released after our LLM knowledge cutoff.\footnote{We used GPT-4o-mini with an October 2023 knowledge cutoff date.}
We conducted refinement experiments identical to Table~\ref{tab:main_result} to correct factual errors in this challenging set.
As shown in Table~\ref{tab:challeng_result}, three rounds of refinement improved \ours by 11.92, comparable to the improvements in Table~\ref{tab:main_result}.
These results confirm our metric provides effective feedback for recent stories independent of LLM parametric knowledge.

\subsection{Latency}
\label{sec:analysis_latency}
\begin{table}[t]
    \centering
    \resizebox{1.0\linewidth}{!}{
    \begin{tabular}{lccc}
        \noalign{\hrule height 1pt}
        \textbf{Metric} 
        & \begin{tabular}[c]{@{}c@{}}\textbf{CKG} \\ \textbf{extraction} \\ \textbf{time}\end{tabular} 
        & \begin{tabular}[c]{@{}c@{}}\textbf{Factuality} \\ \textbf{calculation} \\ \textbf{time}\end{tabular}
        & \textbf{KENDALL} \\
        \hline
        Human              & -    & 132.00 min  & -    \\
        LongDocFACTScore   & -    & 3.81 min & 0.16 \\
        FActScore          & -    & 4.60 min & 0.09 \\
        \ours (w/o consistency) & 0.35 min & 4.77 min & 0.13 \\
        \ours              & 1.17 min & 4.81 min  & \textbf{0.33} \\
        \noalign{\hrule height 1pt}
    \end{tabular}}
    \caption{Average latency (in minutes) per a summary and Kendall’s tau correlation for evaluating factuality across different metrics on the FABLES~\cite{kim2024fables} dataset. Reported total times are the sum of CKG extraction and factuality calculation.}
    \label{tab:latency}
\end{table}

We compared the latency of \textit{LLM-as-a-Judge} metrics, such as LongDocFACTScore and FActScore, with our \textit{Agent-as-a-Judge} metric.
Table~\ref{tab:latency} shows the time required to evaluate the factuality of each summary in the FABLES~\cite{kim2024fables}.
Since long narratives like those in FABLES exceed 100K tokens, human evaluation by verifying details is time-consuming.\footnote{According to \citet{kim2024fables}, annotators spent over 11 hours evaluating five summaries.}
In contrast, LLM-based metrics, including ours, assess factuality within a few minutes.
However, \textit{LLM-as-a-Judge} metrics struggle to assess factuality while understanding character relationships, leading to discrepancies with human evaluations, as shown in Table~\ref{tab:correlation}.
In contrast, \ours devotes additional time to reasoning about character relationships before assessing factuality, resulting in more accurate evaluations despite slightly longer times.

To further analyze efficiency, we also evaluated a one-shot variant of \ours (NFS w/o consistency). 
As reported in Table~\ref{tab:latency}, this variant requires less time for CKG extraction but shows weaker correlation with human judgments, highlighting a trade-off between speed and accuracy.

\subsection{Sensitivity of Metrics on Factual Perturbation}
\begin{table}[]
\centering
\resizebox{1.0\linewidth}{!}{
\begin{tabular}{lccc}
\noalign{\hrule height 1pt}
\textbf{Summary} & \textbf{ROUGE-L} & \textbf{BERTScore$\mathbf{_{f1}}$} & \textbf{NFS} \\
\hline
Reference Summary & 100.00 & 100.00 & 95.42 \\
Perturbed Summary & 81.61 & 92.15 & 40.81 \\
\noalign{\hrule height 1pt}
\end{tabular}}
\caption{Change in metric scores after factual perturbation of the reference summary on the MENSA.}
\label{tab:perturb}
\end{table}
We evaluated metric sensitivity to factual perturbations using GPT-4o with the prompt shown in Figure~\ref{tab:perturb_prompt} on the MENSA. 
Specifically, we perturbed the reference summaries from MENSA by introducing factual inaccuracies in each sentence.
Table~\ref{tab:perturb} shows that ROUGE-L and BERTScore$\mathbf{_{f1}}$ decreased minimally despite factual perturbations, while \ours significantly dropped. 
In our setup, perturbations were created by minimally replacing specific words in the reference summaries rather than generating entirely new content, since the latter would unfairly penalize ROUGE and BERTScore.
These results demonstrate that our metric is highly sensitive to factual discrepancies, making it a suitable metric for assessing factuality.
\section{Conclusion}
This work shows how the agent-as-judge is deployed for narrative summarization to overcome the limitations of existing evaluation metrics, such as overreliance on lexical similarity or factual inconsistencies.
Specifically, we propose consistent CKG extraction, and a new factual evaluation metric based on CKG, and an agent that evaluates and guides the summary and refinement. 
Through our implementation, we demonstrated both the process and superior performance over state-of-the-art methods on real-life industry datasets and scenarios.
\section{Limitation}
We acknowledge two limitations of our work. 
First,  our framework may occasionally retrieve subgraphs unrelated to the atomic fact being evaluated, though this did not impact factuality judgments in our experiments and  outperformed the no-retrieval baseline. Nonetheless, further enhancing subgraph retrieval precision remains a promising direction  for future work.

Second, our framework has been tested exclusively in the narrative domain. Although effective, its generalizability to other domains remains unverified. However, its potential for applications requiring deep character understanding—such as news summarization, biographical writing, and historical analysis—suggests promising directions for future exploration.

\section*{Acknowledgements}
This work was supported by 
the National Research Foundation of Korea (NRF) grant funded by the Korea government (MSIT) (No. RS-2024-00414981),
the MSIT (Ministry of Science and ICT), Korea, under the ITRC (Information Technology Research Center) support program (IITP-2025-2020-0-01789) supervised by the IITP (Institute for Information \& Communications Technology Planning \& Evaluation), and
Institute of Information \& communications Technology Planning \& Evaluation (IITP) grant funded by the Korea government(MSIT) [NO.RS-2021-II211343, Artificial Intelligence Graduate School Program (Seoul National University)].

\bibliography{reference}

\newpage
\appendix
\section*{\centering Appendices}
\section{Further Analysis}
\subsection{Analysis for Baseline Metrics}
\begin{table}[]
\centering
\resizebox{1.0\linewidth}{!}{
\begin{tabular}{lcc}
\noalign{\hrule height 1pt}
\textbf{Strength}             & \textbf{Spearman ($\rho$)}   & \textbf{KENDALL ($\tau$)}    \\
\hline
Very weak correlation   & 0.00\textasciitilde0.15   & 0.00\textasciitilde0.10 \\
Weak correlation          & 0.15\textasciitilde0.30   & 0.10\textasciitilde0.20 \\
Moderate correlation     & 0.30\textasciitilde0.43   & 0.20\textasciitilde0.30 \\
Strong correlation         & 0.43\textasciitilde1.00   & 0.30\textasciitilde1.00 \\
\noalign{\hrule height 1pt}
\end{tabular}}
\caption{Correlation strength based on coefficients.}
\label{tab:coefficient}
\end{table}
Table~\ref{tab:coefficient} shows widely adopted interpretation of correlations from~\citet{botsch2011chapter, chiang-lee-2023-large}, where $|\tau| \in [0.3, 1.0]$ is considered a strong correlation.
For Spearman~\cite{spearman1961proof}, thresholds are derived by converting $\tau$\footnote{$\tau$ indicates Kendall’s tau~\cite{kendall1938new}} intervals under the assumption of bivariate normality.

We analyze the results compared to other metrics in Table~\ref{tab:correlation}.
Metrics based on lexical overlap, such as ROUGE, show stronger correlations with human factuality assessments compared to semantic similarity metrics such as BERTScore, as they better capture repeated entities and locations in narratives.
In contrast, metrics such as BARTScore and LongDocFACTScore \cite{bishop-etal-2024-longdocfactscore}, which rely on log-likelihood and entailment, have lower correlations due to their limited ability to account for broader context and character relationships.
FActScore \cite{min-etal-2023-factscore}, reproduced in our study, incorporates character relationship retrieval to improve factuality assessments.
Building on this, \ours further enhances performance by addressing common errors caused by misinterpreted character relationships, leading to more accurate evaluations.

\subsection{Effectiveness with Open-Source LLM}
\begin{table}[h!]
\centering
\resizebox{1.0\linewidth}{!}{
\begin{tabular}{l|ccccccc}
\noalign{\hrule height 1pt}
\textbf{} & \textbf{R-1} & \textbf{R-2} & \textbf{R-L} & \textbf{BS$\mathbf{_p}$} & \textbf{BS$\mathbf{_r}$} & \textbf{BS$\mathbf{_{f1}}$} & \textbf{\oursshort} \\
\hline
\textbf{\textit{without merging}} & \multicolumn{7}{c}{} \\
~~ TextRank & \textbf{34.37} & 4.60 & 12.84 & 46.86 & 49.43 & 48.10 & 59.72 \\
~~ LED & 17.46 & 1.59 & 10.03 & 42.90 & 42.74 & 42.58 & 56.48 \\
~~ LongT5 & 20.77 & 2.26 & 10.03 & 45.05 & 45.06 & 45.01 & 73.76 \\
\hline
\textbf{\textit{hierarchically merging}} & \multicolumn{7}{c}{} \\
~~ Qwen2.5-14B-Instruct & 30.45 & 7.03 & 11.99 & 58.20 & 59.61 & 58.89 & 67.29 \\
~~ Ours: 1st iteration & 31.26 & 7.06 & 12.05 & 58.31 & 59.68 & 58.95 & 70.71 \\
~~ Ours: 2nd iteration & 31.57 & \textbf{7.08} & \textbf{12.09} & \textbf{58.35} & \textbf{59.69} & \textbf{58.97} & \textbf{71.89} \\
\noalign{\hrule height 1pt}
\end{tabular}
}
\caption{Evaluation results on the MENSA~\cite{saxena2024select} dataset. Hierarchically merging results are based on Qwen2.5-14B-Instruct.}
\label{tab:qwen_result}
\end{table}

Our framework is also effective with open-source model. 
As shown in Table~\ref{tab:qwen_result}, \ours implemented using Qwen2.5-14B-Instruct~\cite{qwen2.5} consistently improves summarization performance on MENSA across two iterations. 
Compared to the initial summarization, our refinement shows steady gains, achieving \oursshort of 70.71 and 71.89 after the first and second iterations, respectively. 
These results confirm that the benefits of our method are not limited to proprietary LLMs but also extend to accessible open-source alternatives.

\subsection{Baseline Evaluation with Hierarchical Merging}
\begin{table*}[h!]
\centering
\resizebox{1.0\textwidth}{!}{
\begin{tabular}{l|ccccccc|ccccccc}
\noalign{\hrule height 1pt} 
\textbf{} & \multicolumn{7}{c|}{\textbf{MENSA}} & \multicolumn{7}{c}{\textbf{MovieSum}} \\
\textbf{} & \textbf{R-1} & \textbf{R-2} & \textbf{R-L} & \textbf{BS$\mathbf{_p}$} & \textbf{BS$\mathbf{_r}$} & \textbf{BS$\mathbf{_{f1}}$} & \textbf{NFS} & \textbf{R-1} & \textbf{R-2} & \textbf{R-L} & \textbf{BS$\mathbf{_p}$} & \textbf{BS$\mathbf{_r}$} & \textbf{BS$\mathbf{_{f1}}$} & \textbf{NFS} \\
\hline
\textbf{\textit{hierarchically merging}} & \multicolumn{7}{c|}{} & \multicolumn{7}{c}{} \\
~~ LED & 22.02 & 3.30 & 9.00 & 43.10 & 43.62 & 43.19 & 59.71 & 21.45 & 2.55 & 14.37 & 44.86 & 44.32 & 44.49 & 61.90 \\
~~ LongT5 & 30.88 & 4.49 & 11.23 & 45.85 & 48.84 & 47.27 & 78.13 & 31.19 & 4.06 & 17.62 & 46.01 & 48.92 & 47.39 & 79.15 \\
~~ GPT-4o-mini & 31.79 & 9.69 & 12.68 & 60.00 & 60.03 & 60.01 & 81.05 & 29.26 & 8.72 & 17.88 & 59.11 & 59.29 & 59.19 & 80.56 \\
~~ Ours: 1st iteration & 33.00 & 9.70 & 12.84 & 60.22 & 60.11 & 60.16 & 85.94 & 30.36 & 8.74 & 18.55 & 59.26 & 59.30 & 59.27 & 86.92 \\
~~ Ours: 2nd iteration & 33.75 & 9.72 & 13.07 & 60.17 & 60.10 & 60.12 & 88.94 & 30.98 & 8.75 & 18.61 & 59.33 & 59.30 & 59.30 & 92.04 \\
~~ Ours: 3rd iteration & \textbf{34.26} & \textbf{9.74} & \textbf{13.46} & \textbf{60.24} & \textbf{60.21} & \textbf{60.22} & \textbf{93.31} & \textbf{31.31} & \textbf{8.81} & \textbf{18.62} & \textbf{59.36} & \textbf{59.31} & \textbf{59.32} & \textbf{94.59} \\
\noalign{\hrule height 1pt}
\end{tabular}
}
\caption{Evaluation results on MENSA and MovieSum datasets with hierarchical merging.}
\label{tab:baseline_merge}
\end{table*}

Our conclusions remain consistent even when baselines are evaluated with hierarchical merging. 
As shown in Table~\ref{tab:baseline_merge}, \ours with refinement continues to outperform LED and LongT5 on both MENSA and MovieSum, confirming the robustness of our improvements regardless of baseline settings.
\section{Qualitative Example}
\begin{table*}[ht!]
\footnotesize
\begin{tabular}{p{1.2cm}p{14.0cm}}
\toprule
\textbf{Initial\newline summary} & 
Saruman reveals to Gandalf that Sauron is regaining strength and is gathering an army to attack Middle-earth, \textcolor{red}{using a Palantir to see his plans}. Gandalf warns Saruman about the dangers of the Palantir and expresses disbelief at Saruman's willingness to ally with Sauron, leading to a fierce battle between the two. In the subsequent scene, \textcolor{red}{Frodo and Sam are enjoying a peaceful moment in the Shire} when they are interrupted by Merry and Pippin, who crash into them after stealing vegetables from Farmer Maggot's field, prompting a humorous chase as they flee from the angry farmer and his dogs. The outcome sees Gandalf and Saruman in conflict, while Frodo and his friends are caught up in a lighthearted escapade. \\
\midrule
\textbf{Atomic\newline facts} & 
[1] Saruman reveals to Gandalf that Sauron is regaining strength. \newline
[2] Sauron is gathering an army to attack Middle-earth. \newline
[3] \textcolor{red}{Sauron uses a Palantir to see his plans.} \newline
[4] Gandalf warns Saruman about the dangers of the Palantir. \newline
[5] Gandalf expresses disbelief at Saruman's willingness to ally with Sauron. \newline
[6] A fierce battle occurs between Gandalf and Saruman. \newline
[7] \textcolor{red}{Frodo and Sam enjoy a peaceful moment in the Shire.} \newline
[8] Merry and Pippin crash into Frodo and Sam. \newline
[9] Merry and Pippin steal vegetables from Farmer Maggot's field. \newline
[10] A humorous chase ensues as they flee from Farmer Maggot and his dogs. \newline
[11] Gandalf and Saruman are in conflict. \newline
[12] Frodo and his friends are caught up in a lighthearted escapade. \\
\midrule
\textbf{Factuality\newline calculation} & 
[1] True \newline
[2] True \newline
[3] False, The statement is false because Sauron does not use a Palantir to see his plans; rather, it is Saruman who uses the Palantir to gain knowledge about Sauron's actions and intentions. \newline
\hspace*{1em} \textit{(evidence scene: \#39, evidence subgraph: `Saruman-own-Palantir', ...)} \newline
[4] True \newline
[5] True \newline
[6] True \newline
[7] False, The statement ``Frodo and Sam enjoy a peaceful moment in the Shire'' is false. The scene depicts Frodo and Sam being interrupted by Merry and Pippin, leading to a messy situation as they are chased by Farmer Maggot and his dogs after stealing vegetables from his field. \newline
\hspace*{1em} \textit{(evidence scene: \#40, evidence subgraph: `Frodo-friend-Sam', ...)} \newline
[8] True \newline
[9] True \newline
[10] True \newline
[11] True \newline
[12] True \\
\midrule
\textbf{Refined\newline summary} & 
Saruman reveals to Gandalf that Sauron is regaining strength and is gathering an army to attack Middle-earth. \textcolor{blue}{Saruman has used a Palantir to gain insight into Sauron's plans.} Gandalf warns Saruman about the dangers of the Palantir and expresses disbelief at Saruman's willingness to ally with Sauron, leading to a fierce battle between the two. In the subsequent scene, \textcolor{blue}{Frodo and Sam are caught in a messy situation in the Shire} when Merry and Pippin crash into them after stealing vegetables from Farmer Maggot's field, prompting a humorous chase as they flee from the angry farmer and his dogs. The outcome sees Gandalf and Saruman in conflict, while Frodo and his friends are caught up in a lighthearted escapade. \\
\bottomrule
\end{tabular}
\caption{Qualitative example illustrating \ours. \textcolor{red}{Red text} in the initial summary and atomic facts indicate factually incorrect statements based on scene evidence, while \textcolor{blue}{blue text} in the refined summary indicate corrections made through agent-based refinement based on feedback.}
\label{tab:qualitative}
\end{table*}

In this section, we illustrate how our approach evaluates factuality, provides actionable feedback, and refines summaries through a qualitative example.
Table~\ref{tab:qualitative} shows the evaluation and refinement process for a summary of \texttt{The Lord of the Rings} generated by GPT-4o-mini~\cite{openai2023gpt4}.
The CKG has already been constructed using the method described in Section~\ref{sec:3_1}.

To evaluate factuality, we decompose the initial summary into atomic facts.
The summary contains two factually incorrect statements highlighted in red, which are also presented in the atomic facts.
First, according to the original script, Saruman uses a Palantir to observe Sauron; however, due to difficulty in understanding character relationships, the incorrect summary stating Sauron observes Saruman was generated and recorded as atomic fact [3].
Second, while the original script depicts Frodo and Sam in a messy situation, chased in the Shire by Merry and Pippin, the summary incorrectly describes it as peaceful, recorded as atomic fact [7].

In our framework, we retrieve relevant scene and subgraph for each atomic fact to evaluate factuality.
Consequently, only [3] and [7] were identified as false.
For these facts, the framework generates not only the factuality but also actionable feedback explaining why they are false and how to correct them.
For [3], based on retrieved scene and relationship that Saruman owns the Palantir, our framework determines that the statement is false and generates feedback suggesting that Saruman uses the Palantir to gain knowledge about Sauron's actions and intentions. 
Similarly, for [7], based on the scene showing Frodo and Sam in a messy situation with Merry and Pippin in the Shire, our framework determines the statement is false and provides proper feedback.
Using this detailed feedback, our framework generates refined summary by correcting only the erroneous parts of the initial summary.

\begin{table*}[ht!]
\footnotesize
\begin{tabular}{p{1.6cm}p{13.6cm}}
\toprule
\textbf{Initial\newline summary} & 
Back at Bag End, Gandalf confronts Frodo about the Ring, revealing its dark history and the resurgence of Sauron, \textcolor{blue}{who seeks it to regain power}. \\
\midrule
\textbf{Atomic\newline facts\newline (1st)} & 
[1] Gandalf confronts Frodo about the Ring at Bag End. \newline
[2] Gandalf reveals the Ring's dark history to Frodo. \newline
[3] \textcolor{blue}{Sauron seeks the Ring to regain power.} \\
\midrule
\textbf{Factuality\newline calculation\newline (1st)} & 
[1] True \newline
[2] True \newline
[3] \textcolor{red}{False,} The statement is judged false because the retrieved subgraph includes the triple \textbf{``Sauron-master of Ring''}, implying that Sauron already possesses the Ring and does not need to seek it. \newline
\hspace*{1em} \textit{(evidence scene: \#29, evidence subgraph: `Sauron-master-Ring', ...)} \\
\midrule
\textbf{Refined\newline summary\newline (1st)} & 
Back at Bag End, Gandalf confronts Frodo about the Ring, revealing its dark history and the resurgence of Sauron, \textcolor{red}{who already possesses the Ring and does not seek it.} \\
\midrule
\textbf{Atomic\newline facts\newline (2nd)} & 
[1] Gandalf confronts Frodo about the Ring at Bag End. \newline
[2] Gandalf reveals the Ring's dark history to Frodo. \newline
[3] \textcolor{red}{Sauron already possesses the Ring and does not seek it.} \\
\midrule
\textbf{Factuality\newline calculation\newline (2nd)} & 
[1] True \newline
[2] True \newline
[3] \textcolor{blue}{True,} Correction: The retrieved subgraph was irrelevant. In fact, Frodo is the Ring-bearer, and Sauron is seeking the Ring to regain power. Thus, the correct statement is that Sauron seeks the Ring. \newline
\hspace*{1em} \textit{(evidence scene: \#25, evidence subgraph: `Frodo-own-Ring', `Sauron-desire-Ring', ...)} \\
\midrule
\textbf{Refined\newline summary\newline (2nd)} & 
Back at Bag End, Gandalf confronts Frodo about the Ring, revealing its dark history and the resurgence of Sauron, \textcolor{blue}{who seeks it to regain power.} \\
\bottomrule
\end{tabular}
\caption{Qualitative example illustrating NarrativeFactScore when irrelevant subgraph is retrieved. \textcolor{blue}{Blue text} in the initial summary and atomic facts indicates the correct fact, but due to irrelevant retrieval, it is first misjudged as false (\textcolor{red}{red}). Through subsequent refinement, the error is corrected.}
\label{tab:qualitative_neg}
\end{table*}

As shown in Table~\ref{tab:qualitative_neg}, this example shows how irrelevant subgraph retrieval can cause errors.  
The atomic fact [3] ``Sauron seeks the Ring to regain power'' is correct.  
However, the retrieved triple ``Sauron-master of Ring'' gives misleading feedback that Sauron already owns the Ring.  
This results in an incorrect refinement, where the fact is wrongly judged as false.  
Through iterative refinement, the system retrieves more relevant subgraphs (e.g., Frodo owns the Ring and Sauron seeks it) or uses scene evidence.  
As a result, the correct factual judgment is recovered in later iterations.

\section{System Deployment}
\begin{figure*}[!t]
 	\centering
 	\includegraphics[width=0.71\linewidth]{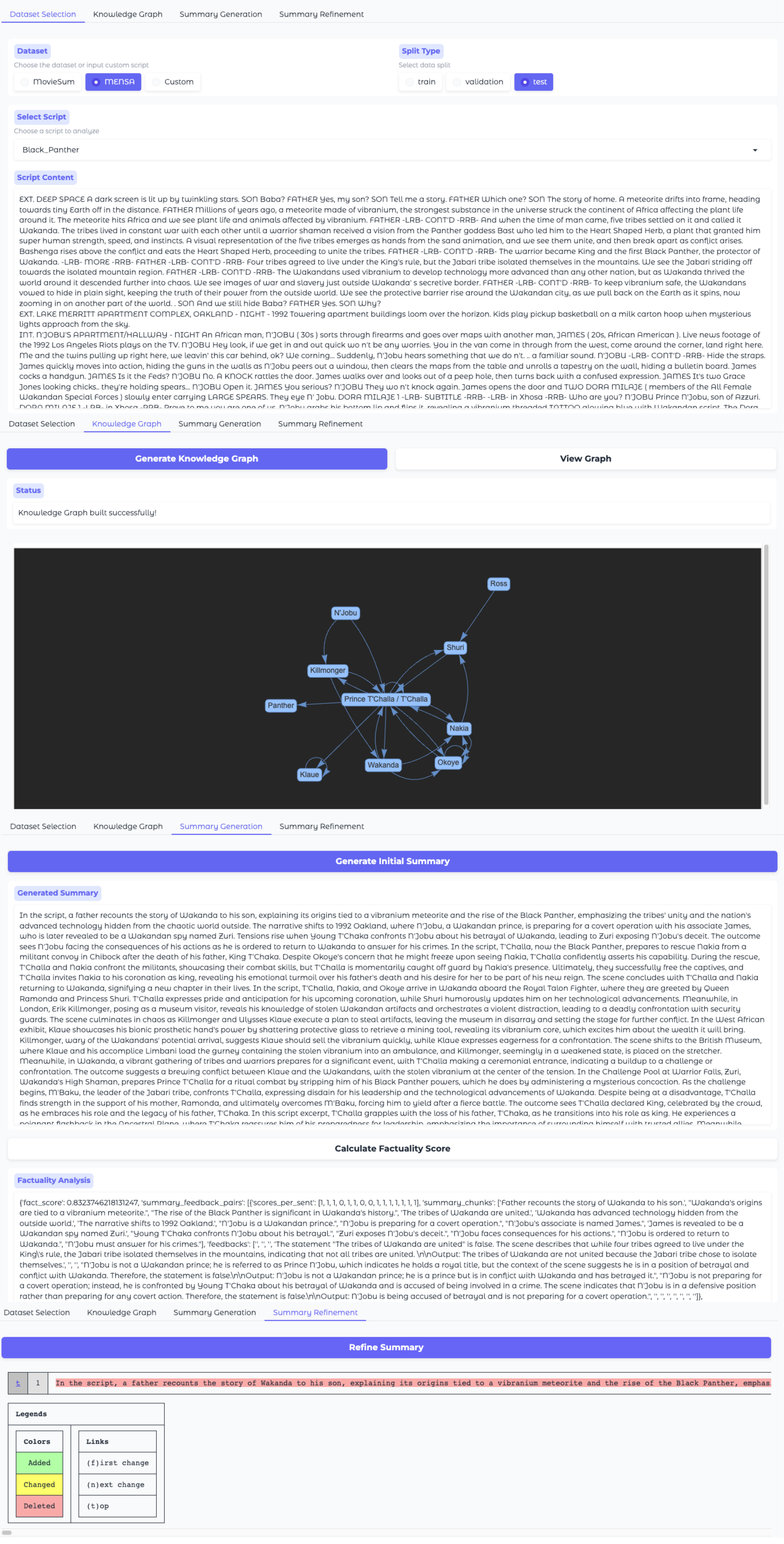}
 	\caption{Deployment overview of \ours.}
 	\label{fig:demo}
\end{figure*}
This section describes our system deployment, which is necessary for the media industry\footnote{The media industry broadly refers to the sector that creates, distributes, and analyzes various forms of narratives, such as movies, television shows, books, video games, and other media that tell stories. This includes businesses involved in producing, editing, and consuming these forms of content, focusing on storytelling in both traditional and digital media.} where companies make investment decisions on narratives (> 100K tokens) such as dramas or movies.
Since reading all narratives is challenging, media companies utilize summaries of each narrative to determine its production feasibility.
However, summaries generated by humans or LLMs frequently contain factual inconsistencies, which hinder accurate investment decisions.
Therefore, our proposed system is deployed to evaluate the factuality of summary for long narratives and improve its factuality.

Figure~\ref{fig:demo} shows screenshots of the system\footnote{\href{https://huggingface.co/spaces/yeonseokjeong/NarrativeFactScore}{huggingface.co/spaces/yeonseokjeong/NarrativeFactScore}}, aligned with the three phases of our framework in Figure~\ref{fig:main_method}.
Using the example \texttt{Black Panther}, users can view the original narrative after selecting a dataset, data type, and name.
Clicking ``Generate Knowledge Graph'' generates and visualizes the CKG (Section~\ref{sec:3_1}). 
The ``Generate Initial Summary'' and ``Calculate Factuality Score'' buttons create an initial summary and evaluate its factuality using the CKG (Section~\ref{sec:3_2}).
Finally, ``Refine Summary'' improves the summary based on feedback, enhancing factuality (Section~\ref{sec:3_3}).
\section{Usage of AI Assistants}
We utilized ChatGPT to improve the clarity and grammatical accuracy of the writing. It provided suggestions for rephrasing sentences and correcting grammatical errors to make the text flow more naturally.
\onecolumn
\clearpage
\section{Prompts}
\label{sec:prompt}
To ensure ethical transparency and reproducibility, we disclose the prompts used at each stage of our process.

\subsection{Knowledge Extraction Prompt for LLM}
\label{sec:prompt_knowledge_extraction}
\begin{tcolorbox}[title=\textbf{Knowledge Extraction Prompt}]
\textbf{[Begin story excerpt]} \\
``Christmas won't be Christmas without any presents,'' grumbled Jo. ``It's so dreadful to be poor!'' sighed Meg, looking out the window at the snow-covered streets of Concord. ``I don't think it's fair...'' \\
... \\
``Glad to find you so merry, my girls,'' said a cheery voice at the door... ``A letter! A letter! Three cheers for Father!'' \\
\textbf{[End story excerpt]} \\

\textbf{Named entities:} \\
Jo / Jo March \\
Meg / Margaret / Margaret March \\
Amy \\
Beth / Elizabeth \\
March sisters \\
Mrs. March / Marmee / Mother \\
Father \\
Concord \\
Union Army \\

\textbf{Knowledge graph edges:} \\
1. Jo, Meg, Amy, Beth; in; March sisters \\
2. March sisters; daughters of; Mrs. March, Father \\
3. Mrs. March; mother of; March sisters \\
... \\
15. Mrs. March; brought home a letter from; Father \\

\textbf{[Begin story excerpt]} \\
\{\textit{scene of narrative}\} \\
\textbf{[End story excerpt]} \\
\end{tcolorbox}
\noindent\begin{minipage}{\textwidth}
\captionof{figure}{Simplified prompt for named entity recognition and knowledge graph edges generation.}\label{tab:named_entity_knowledge_graph}
\end{minipage}

\newpage
\subsection{Narrative Summarization Prompt for LLM}
\label{sec:prompt_initial_summarization}
\begin{tcolorbox}[title=\textbf{Narrative Summarization Prompt}]
This is a part of a script from a Movie. Read the following content carefully, then answer my question: \\
\{\textit{chunk of narrative}\} \\
The script has ended now. \\

\textbf{Summary instructions:} \\
- Provide a detailed summary of the key characters' actions, emotions, and situations as reflected in the dialogue or context. \\
- Clearly state the outcome of the events. \\
- The summary should be between 2 to 5 sentences long. \\
\end{tcolorbox}
\noindent\begin{minipage}{\textwidth}
\captionof{figure}{Prompt for summarizing a chunk of narrative from a movie script.}\label{tab:movie_script_prompt}
\end{minipage}

\subsection{Atomic Fact Decomposition Prompt for LLM}
\label{sec:prompt_fact_decompose}
\begin{tcolorbox}[title=\textbf{Atomic Fact Decomposition Prompt}]
I will give you a summary from a chunk of movie script. \\
Your task is to provide me with a list of atomic facts expressed in the given summary. \\
Each atomic fact should be described in a name-only third-person format. \\
Please separate each atomic fact with a `\textbackslash n`. \\
Summary: \{\textit{sentence of summary}\} \\
\end{tcolorbox}
\noindent\begin{minipage}{\textwidth}
\captionof{figure}{Prompt for extracting atomic facts from a movie script summary.}\label{tab:atomic_fact_extraction}
\end{minipage}

\subsection{Fact-Checking Prompt for \ours}
\label{sec:prompt_fact_check}
\begin{tcolorbox}[title=\textbf{Fact-Checking Prompt}]
Consider the given statement, the related scene, and the relationship subgraph. \\
Indicate whether the statement is supported by the scene and the relationship subgraph. \\
Negation of a false statement should be considered supported. \\
If the statement is true, output 1. \\
If the statement is false, output the reason why it is false. \\
Scene: \{\textit{retrieved scene}\} \\
Relationship Subgraph: \{\textit{retrieved subgraph}\} \\
Statement: \{\textit{atomic fact}\} \\
Output: \\
\end{tcolorbox}
\noindent\begin{minipage}{\textwidth}
\captionof{figure}{Prompt for validating a summary against a scene and a relationship subgraph.}\label{tab:scene_relationship_validation}
\end{minipage}

\newpage
\subsection{Agent-based Refinement Prompt for LLM}
\label{sec:prompt_self_improvement}
\begin{tcolorbox}[title=\textbf{Agent-based Refinement Prompt}]
Below is a part of the script from the titled movie. \\
- Script: \{\textit{chunk of narrative}\} \\
Based on the 'Statement to Revise' and 'Reason for Revision', create a `Revised Summary' of the `Summary of the Script'. \\
Keep the revised summary concise and similar in length to the original summary. \\
Do not directly copy any part of the 'Script.' \\
If the 'Summary of the Script' is accurate, generate the original summary as is. \\
- Summary of the Script: \{\textit{initial summarization}\} \\
- Statement to Revise 1: \{\textit{hallucinated fact atomic}\} (Reason for Revision: \{\textit{feedback}\}) \\
... \\
- Revised Summary: \\
\end{tcolorbox}
\noindent\begin{minipage}{\textwidth}
\captionof{figure}{Prompt for revising and summarizing a movie script based on feedback. Note that `Statement to Revise' and `Reason for Revision' correspond to the atomic fact and factuality feedback calculated in Figure~\ref{tab:scene_relationship_validation}.}\label{tab:script_revision_summarization}
\end{minipage}

\subsection{Factual Perturbation Prompt}
\label{sec:prompt_perturb}
\begin{tcolorbox}[title=\textbf{Narrative Summarization Prompt}]
This sentence serves as a summary of a script. Rewrite this one-sentence summary by minimally replacing a few words in the original sentence to render it factually inaccurate, while keeping the original sentence structure intact. \\

Original sentence: \{\textit{original\_sentence}\} \\
Rewritten sentence:
\end{tcolorbox}
\noindent\begin{minipage}{\textwidth}
\captionof{figure}{Prompt used to generate factual perturbations.}\label{tab:perturb_prompt}
\end{minipage}

\end{document}